\documentclass{article} 
\usepackage{iclr2025_conference,times}

\usepackage{amsmath,amsfonts,bm}

\def\eqref#1{equation~\ref{#1}}

\def\1{\bm{1}}

\DeclareMathAlphabet{\mathsfit}{\encodingdefault}{\sfdefault}{m}{sl}
\SetMathAlphabet{\mathsfit}{bold}{\encodingdefault}{\sfdefault}{bx}{n}

\usepackage{xcolor}
\usepackage{hyperref}
\usepackage{tikz}
\usepackage{url}
\usepackage{amsmath}
\usepackage{amssymb}
\usepackage{graphicx}
\usepackage{array}
\usepackage{algorithm}
\usepackage{algpseudocode}
\usepackage{makecell}
\usepackage{multirow}
\usepackage{wrapfig}
\usepackage{dsfont}
\usepackage{pifont}
\usepackage{authblk}
\usepackage{comment}
\usepackage{verbatimbox}
\usepackage{amsfonts}
\usepackage{textcomp}
\usepackage{fancyvrb}
\usepackage{lettrine}
\usepackage{fancyhdr}
\usepackage{listings}
\usepackage[most]{tcolorbox}
\usepackage{booktabs}
\usepackage{tabularx}
\usepackage{array}
\usepackage{changepage}
\usepackage{float}
\usepackage{geometry}
\newtcblisting{rulebox}[2][]{
  colback=gray!3!white,     
  colframe=gray!50!black,  
  boxrule=0.4pt,           
  arc=2pt,                 
  top=1pt, bottom=1pt,     
  left=2pt, right=2pt,     
  boxsep=1pt,              
  width=\linewidth,     
  listing only,
  title={#2},              
  fonttitle=\bfseries\footnotesize, 
  breakable,
  listing options={
    basicstyle=\ttfamily\footnotesize, 
    keywordstyle=\color{blue}\bfseries,
    commentstyle=\color{gray},
    morekeywords={:-,;,not},
    escapeinside={(*@}{@*)}
  },
  #1
}

\usepackage{enumitem}

\title{Towards Safe Autonomous Driving Policies using A Neuro-Symbolic Deep Reinforcement Learning Approach}

\iclrfinalcopy

\author[1]{Iman Sharifi}
\author[2]{Mustafa Yildirim}
\author[3]{Saber Fallah}
\affil[1]{Dept. of Mechanical and Aerospace Engineering, George Washington University, USA}
\affil[2]{Dept. of Mechanical Engineering, Istanbul University-Cerrahpasa, Istanbul, Türkiye}
\affil[3]{Dept. of Mechanical Engineering Sciences, University of Surrey, UK }
\affil[ ]{\texttt{i.sharifi@gwu.edu, mustafa.yildirim@iuc.edu.tr, s.fallah@surrey.ac.uk}}

\begin{document}
\maketitle





\begin{abstract}
The dynamic nature of driving environments and the presence of diverse road users pose significant challenges for decision-making in autonomous driving. Deep reinforcement learning (DRL) has emerged as a popular approach to tackle this problem. However, the application of existing DRL solutions is mainly confined to simulated environments due to safety concerns, impeding their deployment in real-world. To overcome this limitation, this paper introduces a novel neuro-symbolic model-free DRL approach, called DRL with Symbolic Logic (DRLSL) that combines the strengths of DRL (learning from experience) and symbolic first-order logic (knowledge-driven reasoning) to enable safe learning in real-time interactions of autonomous driving within real environments. This innovative approach provides a means to learn autonomous driving policies by actively engaging with the physical environment while ensuring safety. We have implemented the DRLSL framework in a highway driving scenario using the HighD dataset and demonstrated that our method successfully avoids unsafe actions during both the training and testing phases. Furthermore, our results indicate that DRLSL achieves faster convergence during training and exhibits better generalizability to new highway driving scenarios compared to traditional DRL methods. 
\end{abstract}

\section{Introduction}
Deep reinforcement learning (DRL) is a deep learning technique that involves training an agent to make decisions in an environment by learning from feedback in the form of rewards or penalties \citep{sutton2018reinforcement}. From the perspective of autonomous driving (AD), DRL can be used to enable an autonomous vehicle (AV) to make decisions \citep{li2024lane}, such as lane changes, merging, and maintaining a safe following distance from other vehicles, based on inputs from sensors, such as cameras and lidars. 
One of the advantages of using DRL for AD is its ability to learn from experience. By repeatedly interacting with the environment and receiving feedback, the AV can improve its decision-making ability over time. This makes DRL a promising approach for developing policies for AD that can adapt to the changes in driving conditions and learn from their mistakes. However, a primary challenge of DRL is ensuring the safety of AD, as a safety-critical system, during the exploration phase \citep{gu2022}. In an online setting, exploratory AVs may take actions that lead to disastrous consequences, potentially endangering passengers' lives. Additionally, DRL often requires a large amount of training data \citep{sutton2018reinforcement}, which can be challenging to acquire for AD \citep{kiran2021deep}. These challenges limit the training process of AVs to simulation environments and render their transfer to real-world driving scenarios impractical. Therefore, there is a need for DRL algorithms that can facilitate safe and efficient learning in AD.

Safe DRL techniques \citep{gu2022, garcia2015comprehensive} aim to prevent unsafe outcomes by modifying the optimization process or exploration process \citep{candela2023risk}, e.g. in ADs \citep{yang2023towards, he2022robust, li2022decision, lv2022safe}; one of the optimization-modification approaches is to impose constraints on the expected cost of DRL \citep{achiam2017}, while another approach involves maximizing the satisfiability of safety constraints through a loss function \citep{xu2018semantic}. Additionally, penalties can be introduced to discourage the agent from violating safety constraints \citep{pham2018optlayer,tessler2018reward, memarian2021self}. Alternatively, a more intricate reward structure can be created using temporal logic \citep{de2019foundations, camacho2019ltl, hasanbeig2019reinforcement, jiang2021temporal, den2022planning}. These approaches incorporate all safety concerns into a loss function, making the optimization problem more complicated. Moreover, the knowledge of safety is not distinguishable from the policy \citep{candela2023risk}. 

Exploration modification techniques incorporate external knowledge by using examples provided by a teacher, policies derived from expert demonstrations, or teacher's responses to queries during the learning process \citep{candela2023risk}. In this way, shielding methods \citep{alshiekh2018safe, mazzi2023risk} utilize a shield derived from prior knowledge to directly prevent the agent from taking actions that may potentially lead to safety violations during the exploration of model-based DRL \citep{yang2023safe, jansen2020safe, li2020robust, emam2022safe} and model-free DRL \citep{kimura2021reinforcement}. A shield is a symbolic or non-symbolic logical component that takes the safety constraints into account in order to guarantee safety when exploring an environment \citep{yang2023safe}. Non-symbolic approaches use mathematical expressions to define constraints or rules, while symbolic methods use relational rules that are simple for interpretation and development. Model predictive shielding (MPS) employs a mathematical model of the system to predict the future states and eliminate the actions by which the AV may be in danger \citep{candela2023risk, li2020robust}, but the exact model of the system is not always available, especially in complex, dynamic environments. Similarly, responsibility-sensitive safety (RSS) model \citep{shalev2017formal}, developed by Intel/Mobileye, ensures safety and improves scalability in AD by defining a rigorous mathematical model that considers worst-case scenarios in a way that the AV is not responsible for any accidents \citep{chai2019safety, liu2021calibration}. \cite{makantasis2020deep} extracted ad-hoc safety rules using the RSS model in unrestricted highway driving environments with both autonomous and manual agents, ensuring that AVs will not be responsible for any accidents. \cite{he2023toward} developed an RSS-based safety mask to guarantee the collision safety of the AD agent during both the training and testing processes of reinforcement learning. While used frequently in different scenarios \citep{xu2021calibration, liu2021calibration, zhao2023lc}, it is challenging for RSS-based methodologies to handle intricate scenarios due to mathematical complexities \citep{hasuo2022responsibility}, and there is still room for more interpretability and generalizability to unseen situations. Symbolic logic-driven shielding approaches are commonly used in high-level decision-making tasks to bring transparency and generalization to the next level. \cite{kimura2021reinforcement} employed a shielding method based on logical neural networks (LNNs) \citep{riegel2020logical} to suggest safe actions and avoid useless ones. They have shown that employing external knowledge using LNNs to suggest the proper action list is a promising shielding approach that reduces the training trials while ensuring safety.

Recent advances have explored formal verification frameworks such as Signal Temporal Logic (STL) \citep{arechiga2019specifying} and Differential Dynamic Logic (DDL) \citep{selvaraj2022formal} to ensure safety in autonomous vehicles. For instance, \cite{sahin2020autonomous} propose a hybrid STL and mixed-integer programming (MIP) framework for decision-making and monitoring. \cite{arechiga2019specifying} utilizes STL to specify and verify safety properties of AVs. Similarly, \cite{selvaraj2022formal} apply DDL to formally develop provably safe behaviors in automated driving. \cite{barhoumi2024formal, barhoumi2024formally} investigate traffic safety using conflict-based techniques and adaptive speed control, combining formal logic with real-time traffic data. \cite{nour2023toward} present a framework for monitoring microscopic traffic parameters using STL, further reinforcing the potential of logic-based safety enforcement in dynamic driving environments. Compared to these model-based approaches, our method integrates symbolic first-order logic (FOL) within a model-free DRL framework, offering a rule-driven yet data-adaptive safety filter that is interpretable, flexible, and computationally efficient to implement.

One interesting and effective way to enhance the safety of DRL systems is through injecting human-like reasoning into training process of DRL systems using symbolic logical rules \citep{cropper2022inductive, calegari2020logic}, which are well-suited for integrating human external knowledge \citep{de2019neuro, sarker2021neuro} into the learning process using logical rules \citep{riegel2020logical}. For instance, neuro-symbolic approaches aim to integrate NNs with symbolic reasoning techniques to leverage the statistical learning capabilities of NNs with the interpretability and knowledge-driven reasoning abilities of symbolic methods \citep{dong2019neural}. Symbolic logic provides transparent and interpretable knowledge representation \citep{hazra2023deep}, aiding in understanding and validating the agent's actions. The encoding of domain-specific rules enhances generalization across different tasks, improving the transferability of learned policies and reducing the need for extensive retraining in new scenarios \citep{belle2020symbolic}. Prior knowledge incorporation for guiding the agent reduces sample complexity, leading to data efficiency and faster convergence \citep{kimura2021neuro}. Moreover, symbolic logic enables specifying safety constraints, thereby enforcing the constraints during the learning process and preventing the agent from taking unsafe actions. 

In this paper, similar to the neuro-symbolic approach proposed by \cite{kimura2021reinforcement} and \cite{hazra2023deep}, we aim to enhance the performance of DRL in AD using a state-of-the-art approach to ensure the safety of taken actions during the exploration phase. We introduce a neuro-symbolic framework, called DRL with Symbolic Logic (DRLSL), by incorporating first-order logic (FOL) into DRL, which can be an effective technique in AVs by restricting the action space to enable faster learning and complying with the safety rules. This approach benefits several advantages over previous shielding methods; First, the use of symbolic logic allows for the explicit specification of safety rules, enabling precise control over the agent's behavior in different driving scenarios. The employment of FOL fosters clear-cut reasoning concerning safety constraints and the range of safe actions. This level of transparency simplifies understanding and troubleshooting the system. Second, our method ensures flexibility and adaptability by allowing easy modifications to safety rules without extensive changes to the underlying learning algorithms. This enables quick adjustments to establishing safety requirements and the incorporation of new knowledge. Moreover, the symbolic logic framework refines the action space by eliminating dangerous actions, assisting DRL in finding the optimal policy faster than the traditional DRL. Finally, the DRLSL method enables generalization of learned safe behaviors to new environments. This scalability is crucial for the deployment of AVs in diverse and unpredictable driving conditions.

The main contributions of this paper are as following:
\begin{enumerate}
	\item We introduce a model-free DRLSL framework that integrates symbolic FOL with traditional DRL techniques. This approach ensures the safety of AD systems during the exploration phase of DRL.
	
	\item We compare the proposed method with a vanilla DRL as well as a model-based shielding method and discuss our approach benefits including assured safety, improved learning efficiency, and generalizability in a highway scenario. To improve generalizability, two different highway datasets with different numbers of lanes, velocity ranges, and traffic densities are leveraged, demonstrating that symbolic first-order logic rules can enhance safety and efficiency across diverse environments without requiring post-hoc adjustments\footnote{All data and source codes are available at: \href{https://github.com/CAV-Research-Lab/Safe-Reinforcement-Learning-using-Symbolic-Logical-Programming-for-Autonomous-Highway-Driving}{\texttt{https://github.com/CAV-Research-Lab/DRLSL}}.}.

\end{enumerate}

Overall, we argue that the combination of model-free DRL and symbolic logic can provide a promising avenue for developing safe and reliable AD policies. By leveraging the strengths of both approaches, we can benefit from the adaptability and generalization capabilities of model-free DRL, while also incorporating the explicit rule-based reasoning of symbolic logic. This integration enables the system to handle complex and uncertain driving scenarios, while ensuring compliance with predefined safety rules and constraints.

To the best knowledge of the authors, this is the first time such a hybrid approach has been proposed in the context of autonomous driving. By fusing the power of DRL with logical reasoning, we believe that this novel methodology holds significant potential for addressing the safety and reliability challenges in AVs. Further research and experimentation are needed to explore the full capabilities and limitations of this approach, but early results are promising and suggest that it could pave the way for more robust and trustworthy autonomous driving systems in the future.

The rest of the paper is divided into several sections. Section II introduces the background on DRL. Section III describes the proposed method in general and section IV particularly discusses the method for the AD system. Section V presents the simulation environment, results, and discussion. Finally, section VI draws a conclusion.

\section{Preliminaries}
\subsection{Deep Reinforcement Learning}
Generally, a DRL scenario can be demonstrated using a Markov decision process (MDP) set $(\mathcal{S},\mathcal{A},\mathcal{R},\mathcal{P},\gamma)$ consisting of a set of states, actions, rewards, state transitions, and a discount factor, respectively. At each time step, the agent observes the current state $s \in \mathcal{S}$, selects an action $a$ from the set of available actions $\mathcal{A}$, and receives a reward $\mathcal{R}(s, a, s')$ after transitioning to the next state $s'$. The reward function $\mathcal{R}$ assigns a numerical reward to each state-action pair, which reflects the desirability of that action. The goal of the agent is to learn a policy $\pi(s)$ that maps each state $s$ to an action $a$ in order to maximize the expected cumulative reward. 

Deep $Q$-network (DQN) is a popular model-free DRL algorithm for solving MDPs. It uses a neural network to estimate the $Q$-function, which represents the expected total reward for taking a particular action $a$ in a particular state $s$, and then following the optimal policy thereafter. The $Q$-function is defined as $Q(s,a) = \mathbb{E}[\sum_{t=0}^{\infty} \gamma^t r_t]$, where $\gamma \in [0,1]$ is the discount factor, and $r_t$ is the reward obtained at time step due to action $a$ taken in state $s$. The DQN algorithm uses experience replay and target networks to improve the stability and convergence of the $Q$-function estimates. Specifically, it maintains a replay buffer $\mathcal{D}$ of past experiences, and periodically updates a target network $\hat{Q}$ with the current network $Q$. The agent's loss function is defined as the mean-squared error between the $Q$-values predicted by $Q$ and the target $Q$-values computed using $\hat{Q}$, as shown in the following: 
\begin{equation}
	\begin{aligned}
		L_i(\theta_i) =\: \mathbb{E}_{(s,a,r,s') \sim \mathcal{D}}[(r + \gamma \max_{a'} \hat{Q}(s',a'; \theta_i^-) - Q(s,a;\theta_i))^2] ,
	\end{aligned}
\label{dqn_loss}
\end{equation}
where $\theta_i$ and $\theta_i^-$ are the parameters of $Q$ and $\hat{Q}$, respectively, and the expectation is taken over a random minibatch of experiences from $\mathcal{D}$. Also, $a'$ is the action by which the $\hat{Q}$ value is maximized.
In this scenario, the DQN agent would receive the current state $s_t$ as input and output the $Q$-values for each possible action $a \in \mathcal{A}$. The agent would then select the action with the highest $Q$-value and execute it in the environment.

\subsection{First-Order Logic}

FOL is a knowledge representation and reasoning (KRR) formalism that employs facts and rules to represent knowledge and make logical inferences. In this formalism, a rule consists of a head and a body, following the format: $\verb|head :- body|$,
where \verb|head| denotes an output predicate, expressing a relationship between objects or concepts, and \verb|body| specifies the conditions under which the \verb|head| predicate holds, and \verb|:-| is an entailment operator connecting \verb|body| to \verb|head|, thereby defining if-then rules or clauses. Each predicate comprises a functor and arguments, written as $\verb|functor(arg|_1,\verb|arg|_2,...,\verb|arg|_n\verb|)|$, where arguments can be constants (e.g., specific cars or lanes) or variables (e.g., relative locations and velocities). For example, the predicate $\verb|isInLane(car|_1\verb|, lane|_2\verb|)|$ states that $\verb|car|_1$ is in $\verb|lane|_2$. Similarly, $\verb|safeDistance(car|_1\verb|, car|_2\verb|)|$ may express that $\verb|car|_1$ maintains a safe distance from $\verb|car|_2$. Typically, rules in FOL programs are formulated as Horn clauses, logical implications with a single positive literal\footnote{In logic, a literal is a basic atomic statement that can be either true or false. It represents a single propositional variable or its negation. It is the smallest unit of logic that can be evaluated independently.} (head) and zero or more literals in the body, taking the following form:
\begin{center}
	$\verb|H :- B|_1, \verb|B|_2, ..., \verb|B|_n\verb|.|$
\end{center}
where $\verb|B|_i,i=1,n$ represent premises forming the rule body, and \verb|H|, the rule head, represents a conclusion. This rule implies that if the conjunction of $\verb|B|_1$ to $\verb|B|_n$ is true, then \verb|H| is also true; otherwise, \verb|H| is false. For instance, in the context of autonomous driving, consider the rule:
\begin{rulebox}{}
safeToChangeLane(C):- isInRightLane(C), leftLaneClear(C).
\end{rulebox}
This rule states that it is safe for car $\verb|C|$ to change lanes if it is currently in the right lane and the left lane is clear.

Overall, FOL is suitable for many applications such as autonomous driving due to its expressive power, which allows the specification of complex relational knowledge and safety constraints in an interpretable and modular way. Unlike other formal methods, FOL enables the direct encoding of human-understandable rules and supports deductive reasoning, making it ideal for enforcing safety-critical conditions and enabling transparent decision-making in autonomous vehicles.

\section{Proposed Method: DRLSL}
A symbolic logical program (SLP) employs FOL by manipulating symbolic expressions to represent and reason about knowledge and logical relationships. One of the main benefits of using an SLP is its ability to provide guarantees of reliability and safety. By encoding logical rules and constraints, the system can be designed to obey specific safety rules, such as maintaining a safe following distance or avoiding collisions in AD systems. Additionally, an SLP can enable the system to reason about the consequences of its actions, such as the potential impact of a lane change on other vehicles, and make decisions based on its understanding of the environment. The main goal of using SLP in our proposed DRLSL method is to eliminate unsafe actions from the entire action space in exploration phase for a given state by using symbolic FOL and then give the safe action set ($\mathcal{A}^{safe}_t$) to the DRL agent to ensure safety.

In order to filter unsafe actions, the first step is to define the environment settings and human background knowledge (BK) about the environment. In this context, human BK refers to domain-specific safety rules and commonsense reasoning patterns that humans typically use to assess safe behavior in driving scenarios. This knowledge is formalized as a set of FOL rules that capture relationships between entities (e.g., vehicles, lanes, distances) and define constraints that must hold for actions to be considered safe. To this end, we employ a set of facts that represent the state of the system, including information such as the position, velocity, and acceleration of the agent, as well as those of other objects in the environment. Once the state is defined, the existing logical rules, governed in the environment and known by human beings, can be defined to determine which actions are safe and which are unsafe, assisting in eliminating the unsafe actions. For example, if the agent is in a certain state and there is another object in the environment that is close to the agent, some actions may be unsafe because the agent could collide with the object if those actions are taken. In this case, the rule would define which actions are unsafe based on the relative distance between the agent and the other object.

Once the SLP determines $\mathcal{A}_t^{safe}$ at each time step, the DRL agent employs the $\epsilon$-greedy method to select a safe action $a_t^{safe}$ from $\mathcal{A}_t^{safe}$. This means the agent explores the environment by randomly selecting $a_t^{safe}$ from $\mathcal{A}_t^{safe}$ with probability $\epsilon$, or it selects $a_t^{safe}$ from $\mathcal{A}_t^{safe}$ that maximizes the $Q$-value, with probability $1-\epsilon$. After obtaining $a_t^{safe}$, it is executed by the agent. Subsequently, we calculate the loss function using Eq.~\ref{dqn_loss}, and update the weights of the $Q$-network using the Back Propagation algorithm \citep{lillicrap2020backpropagation}. This process is repeated in different training episodes to find the optimal policy. Referred to as DRLSL, this approach ensures safety during the agent's exploration phase. When deep $Q$-network is employed as a DRL method in this method, it is called DQNSL.

Algorithm~\ref{DQN algorithm} presents the pseudo-code for deep $Q$-network with symbolic logic (DQNSL), which is an instance of DRLSL, and guarantees that the agent only selects actions from the safe action set, preventing the execution of unsafe actions.
\begin{algorithm*}[tb]
	\caption{Deep $Q$-Network with Symbolic Logic (DQNSL)}
	\label{DQN algorithm}
	\begin{algorithmic}[1]
		\State Initialize replay buffer $\mathcal{D}$ with capacity $N$
		\State Initialize action-value function $Q$ with random weights $\theta$
		\State Initialize target action-value function $\hat{Q}$ with weights $\theta^{-} = \theta$
		\State
		Set learning rate $\alpha$ and exploration rate $\epsilon$
		\For{$episode = 1, M$}
		\State Initialize state $s_1$
		\For{$t = 1, T$}
		\State SLP extracts $\mathcal{A}^{safe}_t$ in $s_t$ \hspace{0.2cm} $\rightarrow$ (Section~\ref{SLPforAD})
		\State With probability $\epsilon$, select a random action from $\mathcal{A}^{safe}_t$ as $a_t^{safe}$ 
            \State Otherwise, select $a_t^{safe} = \max_{a} Q(s_t, a|\mathcal{A}^{safe}_t; \theta)$
		\State Execute action $a_t^{safe}$ and observe reward $r_t$ and next state $s_{t+1}$
		\State Store transition $(s_t, a_t^{safe}, r_t, s_{t+1})$ in $\mathcal{D}$
		\State Sample random minibatch of transitions $(s_j, a_j^{safe}, r_j, s_{j+1})$ from $\mathcal{D}$
		\State Set target value for minibatch transition $j$:
		\begin{equation*}
			y_j = 
			\begin{cases}
				r_j & \text{for terminal } s_{j+1} \\
				r_j + \gamma \max_{a'} \hat{Q}(s_{j+1}, a'; \theta^{-}) & \text{for non-terminal } s_{j+1}
			\end{cases}
		\end{equation*}
		\State Update $Q$ by minimizing the loss:
		\begin{equation*}
			\mathcal{L}(\theta) = \frac{1}{N} \sum_{j}(y_j - Q(s_j, a_j; \theta))^2
		\end{equation*}
		\State Every $C$ steps, update the target network: $\theta^{-} \leftarrow \theta$
		\State Decrease $\alpha$ and $\epsilon$ linearly over time until reaching the minimum values
		\EndFor
		\EndFor
	\end{algorithmic} 
\end{algorithm*}

\section{Safe Learning of Autonomous Driving}
In this section, we implement the proposed DRLSL method on the AD system to show its superior performance and effectiveness on safe exploration. An SLP can be used to ensure the safety of the AD system by encoding FOL-based rules and constraints related to driving behavior. For example, the system can be designed to follow traffic rules, maintain a safe following distance, or avoid collisions. Also, an SLP can be integrated with DRL techniques by incorporating logical driving constraints into the policy optimization process of the DRL agent. By incorporating safety rules as constraints, the system can be trained to optimize its behavior while still obeying the rules. 

\subsection{Highway Driving Environment}
In this section, we define the states, actions, and reward function for the AD system to prepare the environment for the DRLSL agent.

\begin{figure*}
    \label{dist1}
    \centerline{\includegraphics[width=0.5\linewidth]{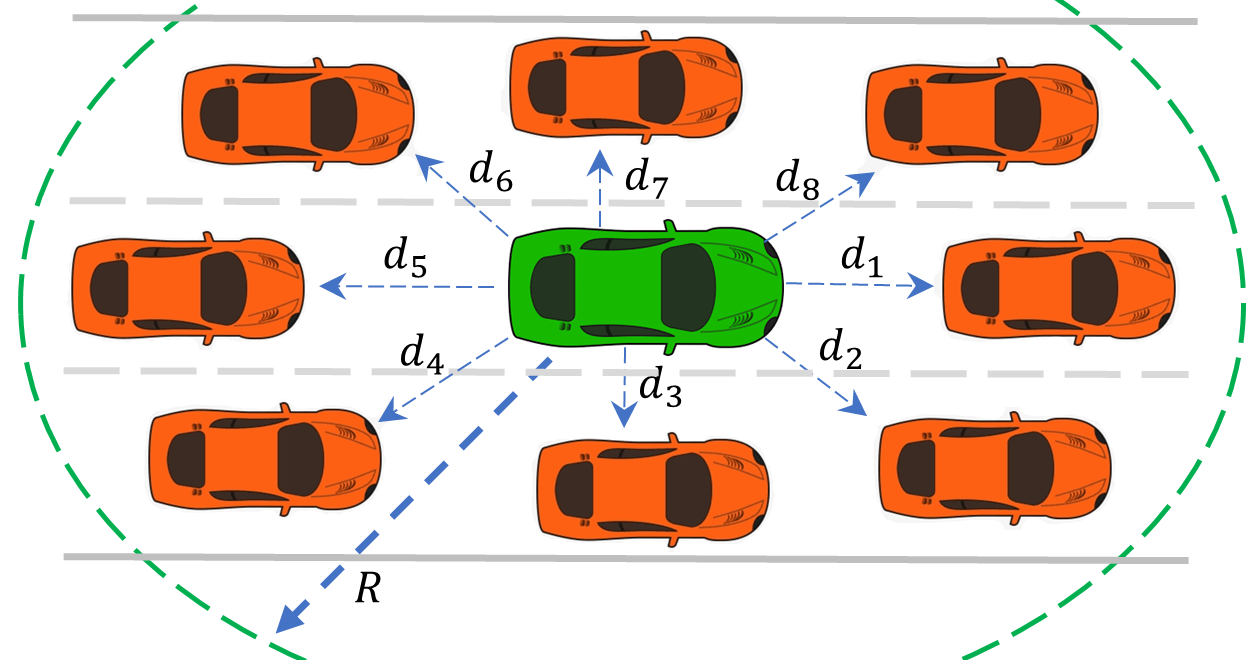}}
    \caption{The distances between the AV (Green vehicle) and each TV (Red vehicles)}
    \vspace{-0.0in}
\end{figure*}

\begin{figure*}[htbp]
	\centerline{\includegraphics[width=0.8\textwidth, scale=0.5]{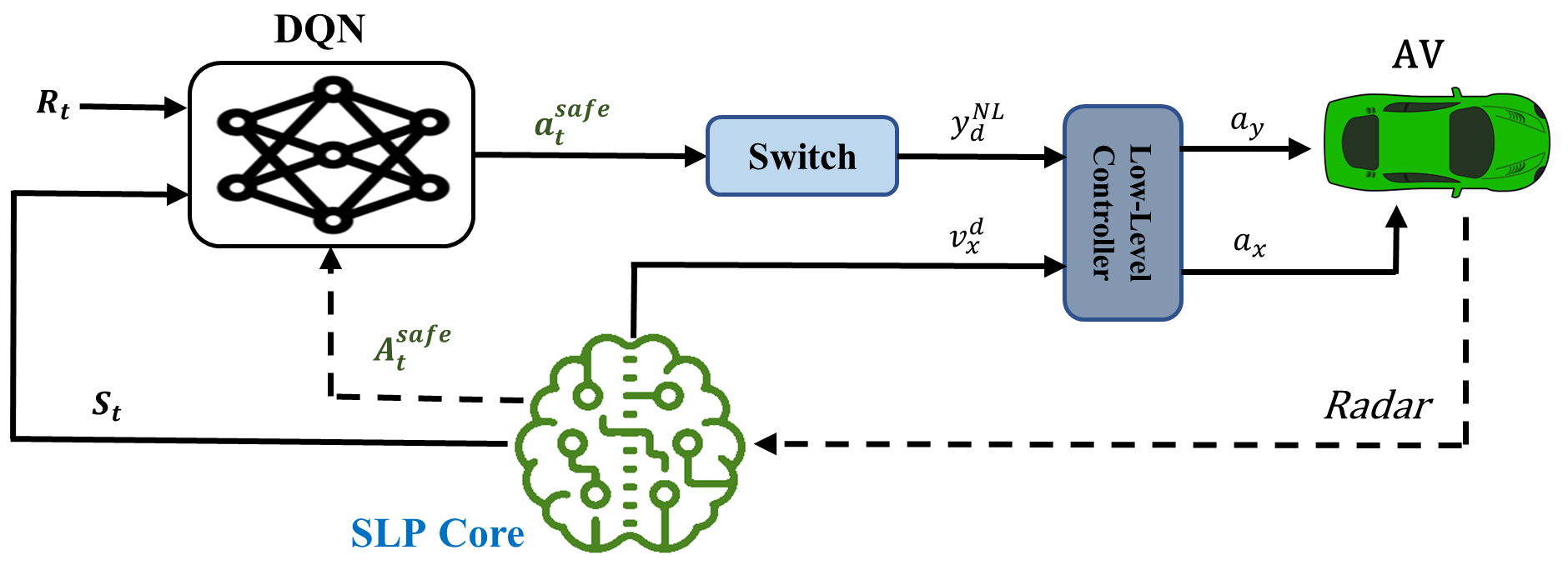}}
	\caption{Deep Q-Network with Symbolic Logic (DQNSL) using a Symbolic Logical Program (SLP)}
	\label{DQNSL}
\end{figure*}

\textbf{State Space. }
The state space $\mathcal{S}$ consists of the distances between the AV and each target vehicle (TV), including $\{d_1,d_2,...,d_8\}$, as shown in Figure~\ref{dist1}, and also the AV longitudinal velocity ($v_x$). After computing the required parameters for the state of the system, they are normalized between $0$ and $1$. This is the data pre-processing step and required for better convergence of the DRL network. To normalize the distance between the AV and TVs, we divide each distance by the radar range $R$, as shown in Figure~\ref{dist1}. Also, we divide the velocity of the AV by the maximum value in each time step. Thus, the state can be represented as $s_t = \{\bar{d_1},\bar{d_2},...,\bar{d_8},\bar{l},\bar{v}_x\}$, where $\bar{d_i}=\frac{d_i}{R}$  and $\bar{v}_x=\frac{v_x}{v_{max}}$. $v_{max}$ is the maximum the maximum allowed speed in the highway. We extract the state, from the surrounding vehicles information, using the logical rules defined in the SLP core, as shown in Figure~\ref{DQNSL}.

\textbf{Action Space. }
The action space $\mathcal{A}$ includes the set of possible actions that the AV agent can take. Since the main goal is to avoid collisions by efficiently changing the lane, the action space includes three symbolic actions, called \verb|lane_keeping|, \verb|left_lane_change|, and \verb|right_lane_change| which are converted to [0, 1, 2] in the DRL neural network, respectively. Unlike \citep{baheri2020deep,wang2019lane}, $v_x$ is removed from the action space since we employed a rule-based longitudinal velocity scheme which adjusts the longitudinal acceleration based on the availability and distance of the front vehicle. The associated rules and parameters are designed in a way producing accelerations with small values, thereby creating a smooth driving. This scheme is discussed in-depth in the Longitudinal Velocity Control sub-section.

\textbf{Reward Function. }
We employ a pre-defined reward function to help DRL to find the optimal policy. Reward function exerts a big impact on the training of DRL policy. Before introducing the reward structure, let's assume the agent moves from the state $s$ to the next state $s'$ by taking action $a$ at a given time step. The reward function for the AV is defined as follows:
\begin{equation}
	\begin{aligned}
		R(s,a,s') = \; w_{lc} \: r_{lc}(s,a) + w_v \: r_{v}(s, a) 
		 + w_c \: r_{c}(s,a) + w_{out}\: r_{out}(s, a) ,
	\end{aligned}
\end{equation}
where $r_{lc}(s,a)$, $r_v(s,a)$, $r_c(s,a)$, and $r_{out}(s,a)$ are the lane-change reward, velocity reward, collision reward, and off-road reward, respectively, and $w_{lc}$, $w_v$, $w_c$, and $w_{out}$ are the correspondent weights. To take the safety of passengers into account, the smoothness of the AV's path on the road should be high. Therefore, the AV should not change the lane except for necessary times. That is why we devised the following lane-change reward:
\begin{equation}
	\begin{aligned}
		r_{l}(s, a) = \begin{cases}
			-1 & \text{if AV changed the lane} \\
			0 & \text{otherwise.} \\
		\end{cases} 
	\end{aligned}
\end{equation}

To make the agent agile, we must consider a velocity reward. We devised a reward function for the AV's velocity in order to drive in the lanes with maximum allowed speed. Since we control the velocity of the AV using the rule-based methods, the AV does not have any control over the velocity. However, it can change the lane to find the maximum allowed velocity. Generally, the velocity reward is equal to $v_x$ with a correspondent weight, as $r_v(s,a)=v_x$.
To avoid collision, the collision reward is defined as follows:
\begin{equation}
	\begin{aligned}
		r_{c}(s, a) = \begin{cases}
			-(1-\eta_c\frac{x}{X}) & \text{if collision occurred} \\
			0 & \text{otherwise,} \\
		\end{cases}
	\end{aligned}
\end{equation}
where $x$ is the longitudinal position of the point where a collision happens and $X$ is the total road length. $\eta_c \in [0,1]$ is a collision factor that penalizes agents for early collisions, preventing them from earning high rewards despite short participation. It encourages prolonged collision avoidance and can also serve as a normalization term within each episode. The parameter allows tuning the collision penalty relative to $x$ and other reward components. The AV must not drive out of the legal lanes of the highway. Thus, as same as the collision reward, the lane deviation penalty reward function is given by:
\begin{equation}
	\begin{aligned}
		r_{out}(s, a) = \begin{cases}
			-(1-\eta_o\frac{x}{X}) & \text{if AV drives off the road} \\
			0 & \text{otherwise,} \\
		\end{cases}
	\end{aligned}
\end{equation}
where $\eta_o \in [0,1]$ plays a similar role in the lane deviation reward as $\eta_c$ does in the collision reward. In our experiments, we set $\eta_c = \eta_o = 0.8$.
We have considered the weights for each sub-reward in a way balancing the magnitude of each sub-reward during a given training episode. Using this technique, the contribution of each sub-reward in the total reward is nearly equal. Accordingly, we assigned $5$, $0.01$, $100$, and $100$ to $w_{lc}$, $w_v$, $w_c$, and $w_{out}$, respectively.

After defining the environment components, we can define the governing rules for the highway driving to generate $\mathcal{A}^{safe}_t$, as presented in the following subsection.

\subsection{SLP for AD}
\label{SLPforAD}
In order to extract $\mathcal{A}^{safe}_t$ for the AV in each step, an SLP uses all important information of the AV and TVs which received by the installed radar on the AV. We employ Prolog \citep{korner2022fifty}, as a symbolic logical programming language to define the desired safety rules in a highway environment. 

To begin with, the useful data for the main criterion includes vehicles' IDs, lanes, dimensions, as well as longitudinal and lateral positions and velocities. Vehicles' data is saved in each step and sent to the SLP core with the following fact clause:
\begin{rulebox}{}
safe_actions(Action):-
            right_is_valid, right_is_safe, Action = right_lane_change.
\end{rulebox}
where \verb|ID| shows the identity code of a given vehicle, \verb|Lane| is the vehicle lane, $\verb|(P|_x\verb|,P|_y\verb|)|$ are the longitudinal and lateral positions, $\verb|(V|_x\verb|,V|_y\verb|)|$ are the longitudinal and lateral velocities, and \verb|(W,H)| are the length and width of the vehicle, respectively. 

Afterward, we utilize human BK to shield unsafe actions from $\mathcal{A}$. Finding the target vehicles using the installed radar on the AV, the relative positions of the TVs respect to the AV are described using their distance and the lanes. To do so, we divide the area around the AV to eight distinct sections; \verb|front|, \verb|front_right|, \verb|right|, \verb|back_right|, \verb|back|, \verb|back_left|, \verb|left|, and \verb|front_left|. This will assist us in finding the relative positions of each target vehicle in respect to the AV location. Knowledge of the relative positions of each TV helps us to find the busy sections around the AV, thereby making the decision-making process straight-forward. As an example, let's consider this rule, "If there is a TV in front of the AV and it has a similar lane to the AV lane, then the front of the AV is busy (\verb|front_is_busy|)." We can show this rule using the following rule clause:
\begin{rulebox}{}
front_is_busy:-
  lane(ego, Lane1), position_x(ego, Px1), target_vehicle(Car),
  lane(Car, Lane2), position$_x$(Car, Px2), Lane1 is Lane2,
  direction(ego, Direction), (Direction = right_to_left, Px2 < Px1;
  Direction = left_to_right, Px2 > Px1), distance(ego, Car, D), D>0.
\end{rulebox}

where \verb|ego| and \verb|Car| represent the AV's and front TV's ID, $\verb|Lane|_1$ and $\verb|Lane|_2$ indicate their lanes, and $\verb|P|_x^1$ and $\verb|P|_x^2$ indicate their longitudinal positions, respectively. Predicate \verb|target_vehicle(Car)| identifies the TVs using the euclidean distance from the AV, and the AV identifies the front TV using the relative longitudinal position in each direction. Thus, \verb|front_is_busy| is a predicate with boolean values to show when the front of the AV is busy. If a location is busy, then it is not free, as indicated in  \verb|front_is_free:- not(front_is_busy)|. 
Similarly, we repeat the procedure for the rest of the sections around the AV to find the busy ones. When the AV knows about the busy and free sections, it can change the lane to the free lanes and avoid busy ones. 

In the next step, a set of logical rules is applied to determine $\mathcal{A}^{safe}_t$. For example, the rule "if there is no vehicle in the left sections, either keep your lane with a safe velocity or go to the left one," or "if the right and left sections are busy, then keep the lane" can be used to extract $\mathcal{A}^{safe}_t$ for lane changes. 
To implement the safety module, we have written an FOL program that consists of a set of rules that encode safety constraints. The following rule encodes the constraint that \verb|left_lane_change| is safe only if going to the left section is valid (\verb|left_is_valid|) and is safe (\verb|left_is_safe|) at the same time, as shown in the following rule:
\begin{rulebox}{}
safe_actions(Action):- 
            left_is_valid, left_is_safe, Action = left_lane_change.
\end{rulebox}
The predicate \verb|left_is_valid| is unified with one if and only if the AV does not drive off the road by choosing the \verb|left_lane_change| action. Likewise, the \verb|left_is_safe| predicate is true when there is no TV in the left sections and the AV does not involve in collisions by taking the \verb|left_lane_change| action. 
Similarly, if there is no vehicle in the right side of the AV (\verb|right_is_safe|) and the right sections are valid (\verb|right_is_valid|), then the \verb|right_lane_change| action is safe, as shown in the following rule:
\begin{rulebox}{}
safe_actions(Action):-
            right_is_valid, right_is_safe, Action = right_lane_change.
\end{rulebox}
Furthermore, \verb|lane_keeping| is always available for the autonomous vehicle to take. Since we design a velocity control scheme for the autonomous vehicle using an SLP in the next sections, the AV can take \verb|lane_keeping| in each state and ensure avoiding collision with the front vehicle. 

After generating all required rules, if the predicate \verb|safe_actions/1| succeeds, it means that the taken action is safe, and the \verb|Action| variable is unified with the corresponding action. The \verb|safe_actions| predicate returns each $a^{safe}_t$ separately, but in order to find $\mathcal{A}^{safe}_t$, we need to backtrack \citep{ciatto2021lazy} all individual safety rules and save their outputs to a list. This process can be performed using a built-in predicate named \verb|findall/3| which takes three arguments, including the \verb|Action| variable, the \verb|safe_actions| functor, and an arbitrary list name like \verb|Actions|. Using the input arguments, this predicate will find all of the safe actions individually and save them to the \verb|Actions| list. 

\subsection{DRLSL for AD}
As shown in Figure~\ref{DQNSL}, once $\mathcal{A}^{safe}_t$ is determined using the FOL-based rules in the SLP core, we use PySwip library, as an interface between Prolog and Python, to connect the SLP core to the DRL agent. Thus, $\mathcal{A}^{safe}_t$ is passed to the DQN network as the available actions for the current state. When DQN agent employs symbolic logic (DQNSL), it chooses the best action from $\mathcal{A}^{safe}_t$ based on the $\epsilon$-greedy method. By restricting the entire action space to the $\mathcal{A}^{safe}_t$, we ensure that the DQN network learns only the safe actions and comply with the safety rules. 

Integrating the SLP into DRL can lead to DRLSL, which considers safety constraints in addition to the policy of maximizing reward. The advantages of this approach include the ability to extract safe actions, which can reduce the risk of accidents and increase the reliability of the AD system. An SLP can also provide a transparent and interpretable framework for reasoning about safety, making it easier to verify and certify the system.

\begin{figure*}[htbp]
	\centerline{\includegraphics[width=\textwidth]{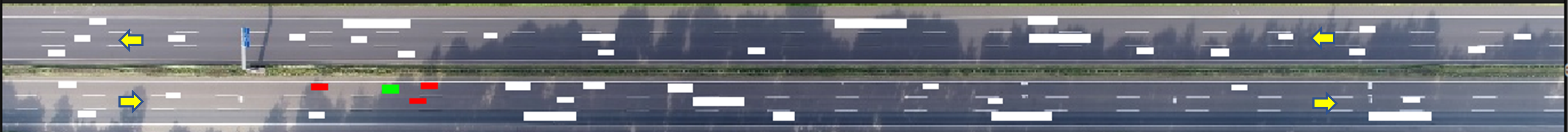}}
	\caption{Highway driving environment in Pygame (left-to-right direction). Green, Red, and White rectangles are AV, TVs, and other vehicles on the road, respectively.}
	\label{vehicles}
\end{figure*}

It is worth noting that after finding $\mathcal{A}^{safe}_t$, DQNSL picks $a^{safe}_t$ from it. According to the taken action and the would-be lane, we find the lateral center of the next lane ($y^{NL}_d$). The process of finding the desired lateral center of each lane is performed using a Switch box, as indicated in Figure~\ref{DQNSL}. According to Figure~\ref{vehicles}, for example, assume the AV is driving in the left-to-right direction and the middle lane (the fifth lane from top) and DQNSL commands to take the \verb|right_lane_change| action. Thus, the sixth lane is the next lane, meaning that the $y^{NL}_d$ is $y^6_d$. Taking the $y^6_d$ as the desired output, a Proportional-Integral (PI) controller, as a low-level controller, produces a lateral acceleration $a_y$ to control the lateral position $y$ of the AV, as shown in Figure~\ref{DQNSL}.

\section{Performance Evaluation}
In this section, firstly, we describe the simulation settings and DRLSL agent's network structure. Afterward, we discuss the longitudinal velocity control scheme. Finally, we elaborate upon the implementation of the proposed method and discuss the results.

\subsection{Dataset}
Numerous open-source platforms, including Carla \citep{dosovitskiy2017carla} and AirSim \citep{shah2018airsim}, exist to simulate traffic environments. However, it was difficult to implement simulated traffic based on actual data on these platforms. Our goal was to simulate a real-world agent by predicting drivers' intentions, which made these platforms inappropriate. As a result, we created a Pygame-based traffic simulation environment \citep{mcgugan2007beginning}, as shown in Figure~\ref{vehicles}.

Several studies \citep{ji2023hierarchical,chen2023safe,tang2022prediction} have examined highway driving by utilizing the NGSim \citep{yeo2008oversaturated} and HighD \citep{krajewski2018highd} datasets. Both collections provide information regarding the lateral and longitudinal positions, velocities, and accelerations of each vehicle on the road. The HighD dataset, in particular, is filmed using a drone camera that captures a $420$-meter section of a German highway, offering accurate data even in situations where traffic is obstructed. The current research relies on the HighD dataset for both training and testing, using one of the most dense tracks for each stage of the study from a total of $60$ tracks in the dataset, as depicted in Figure~\ref{vehicles}. Although this method of studying traffic provides realistic results, it carries some limitations. Since we add a virtual AV to learn to drive efficiently among other vehicles, those vehicles are not aware of the AV's presence. In some situations, this leads to unavoidable accidents of whom the AV is not the cause. For example, when the AV is driving in a lane, another vehicle collides with it from back since the vehicle does not know the AV is on the road.

According to the HighD dataset, we assume an AV is driving on a two-directional, three-lane highway, as shown in Figure~\ref{vehicles}. To clarify the position of each vehicle, we assign $1$ to $6$ to the highway lanes from top to bottom of Figure~\ref{vehicles}. In the first three lanes, vehicles move in the right-to-left direction and in the rest, vehicles move in the left-to-right direction. 

\subsection{Network and Hyperparameters}
Initially, we conducted assessments to identify hyperparameters that would ensure optimal and swift convergence. We determined the parameters after running preliminary tests. The network consisted of three fully connected layers, with the first two layers consisting of $256$ nodes and the final layer consisting of three nodes to represent the action space. We employed the PyTorch library for neural network computations. Moreover, the ADAM algorithm, which is computationally efficient and rapidly converges, was used to optimize the networks. The discount factor was assigned a value of $\gamma=0.95$, while the learning rate was initially set as $\alpha_i=0.01$ and decays over each episode to reach the final value $\alpha_e=1e-4$. The experience replay memory $\mathcal{D}$ size is $100,000$, and stochastic gradient descent batch size is 128. Except for the noisy networks, an epsilon-greedy exploration policy was implemented with a starting value of $\epsilon_i=0.1$ and decreasing to a minimum value of $\epsilon_e=0.001$. To stabilize the Q-network in the case of convergence, we updated the target network after 1,000 iterations. After describing the simulation settings, the defined reward function and the way we control $v_x$ to avoid collision with the front vehicle will be taken into account.

\subsection{Baselines: DQN and MPSDQN}
To compare the results of the proposed method, we implement a vanilla DQN and safe Model Predictive Shielding (MPS) \citep{bastani2021safe, li2020robust} with DQN (MPSDQN). Generally, MPS methods are safe model-based paradigms assuming an accurate model of the system is available all the time. Using the model, they predict next states of the system by taking different actions in the current state, and by evaluating the safety of next states, they can extract safe actions and eliminate unsafe ones. However, systems modeling is not always trivial due to disturbances or humans' free will. Here, using the HighD dataset, positions and velocities in each state are used to predict the next state of other vehicles via a simple linear model for each vehicle. After achieving next states of the AV and TVs, we identify whether a collision happened or the AV went off-road. Using so, we rule out the corresponding actions from the action space. 
Having extracted the safe action set, a DQN agent tries to find the best action among the set. The DQN network parameters are similar for all agents.

\subsection{Longitudinal Velocity Control}
In this section, we explain how the desired velocity of the AV is obtained in each state. We have written FOL-based rules for determining the longitudinal acceleration $a_x$ of the AV, which takes into account the current lane, maximum speed limit of the lane, the distance and speed of the front TV. The goals of the designed rules are collision avoidance with the front TV, making the driving smooth, as well as reducing the duration of driving along the track in the highway. 
The rules consider three scenarios; firstly, if there is no vehicle in front of the AV (\verb|front_is_free|), then the desired velocity is set to the maximum speed limit of the current lane, reducing the time needed for moving along the highway and, at the same time, obeying the speed limit rules. Accordingly, $a_x$ is defined as following:
\begin{equation}
	a_x = \frac{v_{max}-v_{x,AV}}{\Delta t},
\end{equation}
where $\Delta t$, $v_{max}$, and $v_{x,AV}$ are the time step, the maximum allowed velocity in a given lane and the current velocity of the AV, respectively. As the second scenario, if there is a vehicle in front of the AV (\verb|front_is_busy|), the desired velocity is calculated based on the distance and speed of that vehicle. In this case, we use a critical distance (which can be computed using RSS models for minimum distances) and actual distance to calculate the acceleration of the AV. If the distance is greater than the critical distance, $a_x$ is calculated using the following equation:
\begin{equation}
	a_x = \frac{v^2_{x,TV}-v^2_{x,AV}}{2(D-C)},
\end{equation}
where $v_{x,TV}$ is the longitudinal velocity of the front TV. Also, $D$ and $C$ are the actual distance and the critical distance between the AV and the TV. Here, we assume that $D$ is bigger than $C$. As the third scenario, when $D$ is lower than $C$, the AV must decelerate to avoid collision with the front TV, as shown in the following equation:
\begin{equation}
	a_x = \frac{-v^2_{x,AV}}{2D}.
\end{equation}

After obtaining the desired acceleration, we update the desired velocity of the AV using $v_x^d=v_x+a_x\Delta t$ and after validating the velocity magnitude, it will be sent to a low-level controller, as shown in Figure~\ref{DQNSL}. Indeed, if the calculated velocity is greater than the maximum speed limit of the lane, the desired velocity is set to the maximum speed limit. Therefore, we make sure the AV follows the speed limit rule. Using a PI controller, as a low-level controller, helps to produce a continuous and smooth $a_x$ and avoid collision with the front vehicle.
\begin{figure*}[t]
	\centering
	\begin{minipage}[b]{\textwidth}
		\centering
		\includegraphics[width=0.8\textwidth]{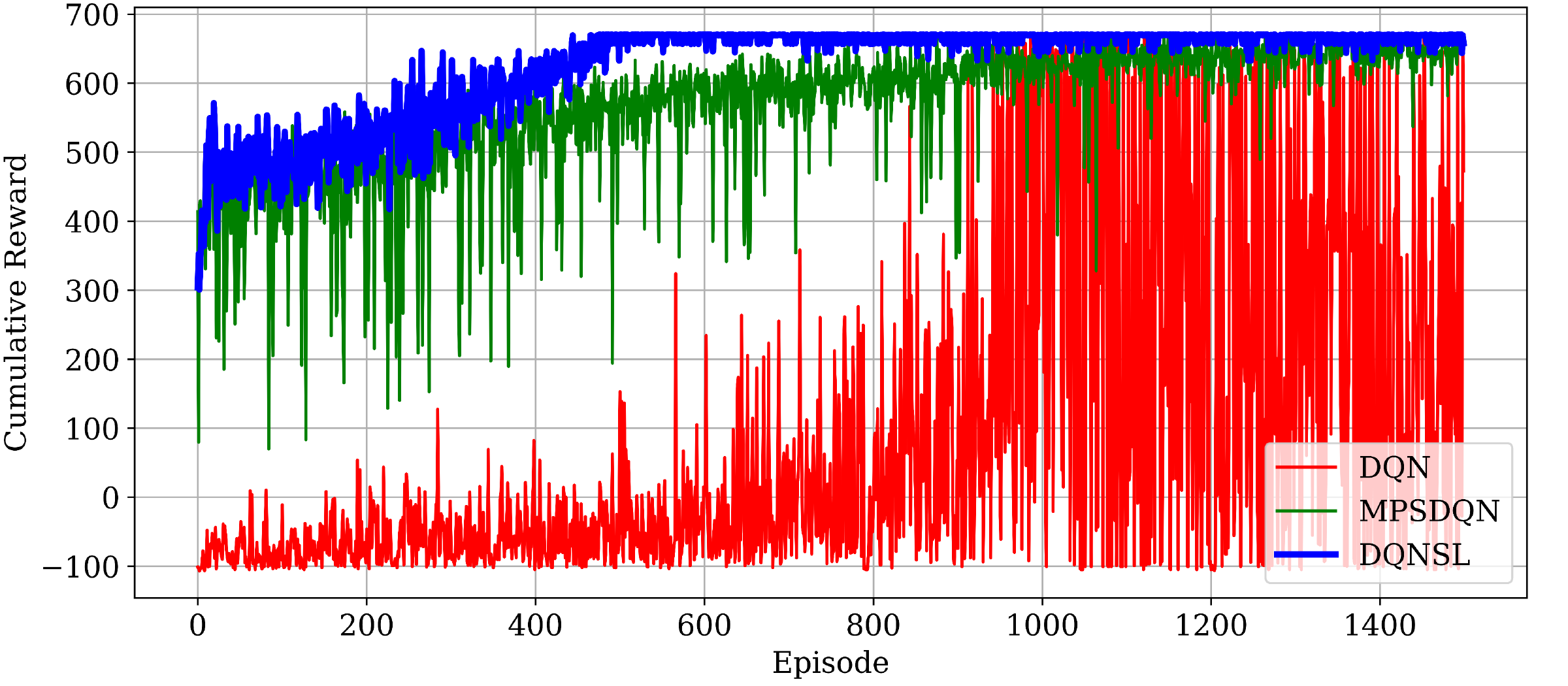}
		\caption{Cumulative reward received by agents in each episode}
		\label{cumulative}
	\end{minipage}
	\hfill
        \vspace{0.2cm} 
        \begin{minipage}[t]{\textwidth}
		\centering
		\begin{minipage}[b]{0.48\textwidth}
			\centering
			\includegraphics[width=\linewidth]{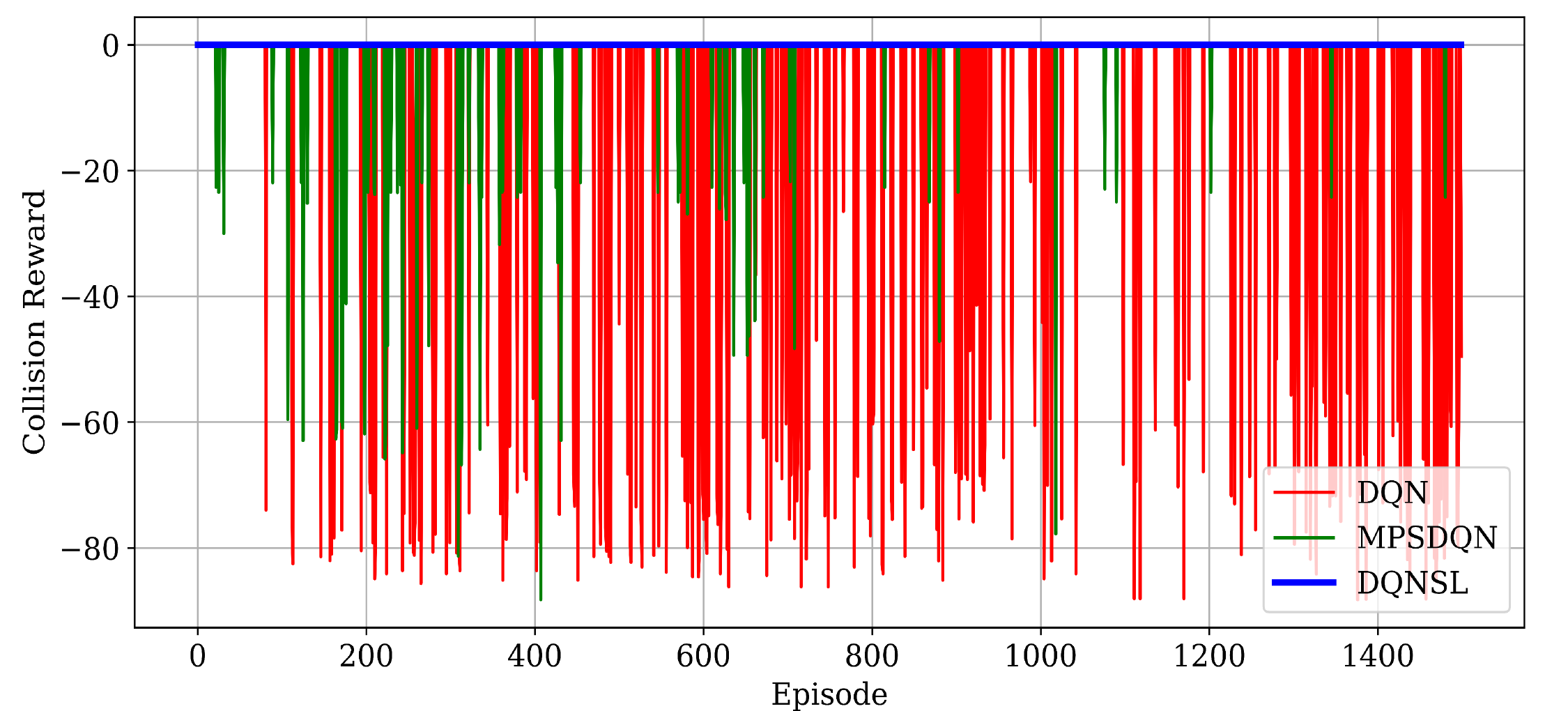}
			\caption{Collision rewards received by agents in each episode}
			\label{collision}
		\end{minipage}
		\hfill
		\begin{minipage}[b]{0.48\textwidth}
			\centering
			\includegraphics[width=\linewidth]{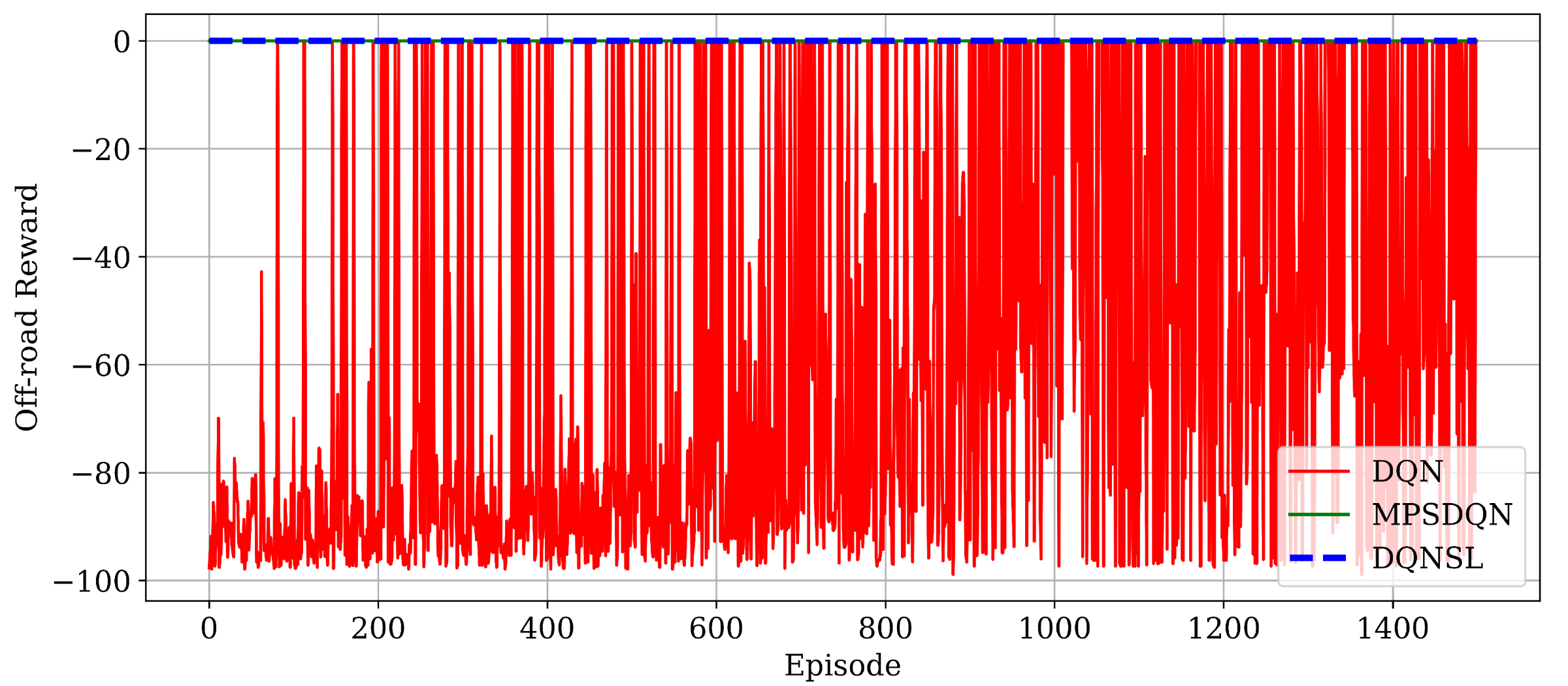}
			\caption{Off-road rewards received by agents in each episode}
			\label{outofhighway}
		\end{minipage}
	\end{minipage}
	\vspace{0.2cm}
	\begin{minipage}[t]{\textwidth}
		\begin{minipage}[b]{0.48\textwidth}
			\centering
			\includegraphics[width=\textwidth]{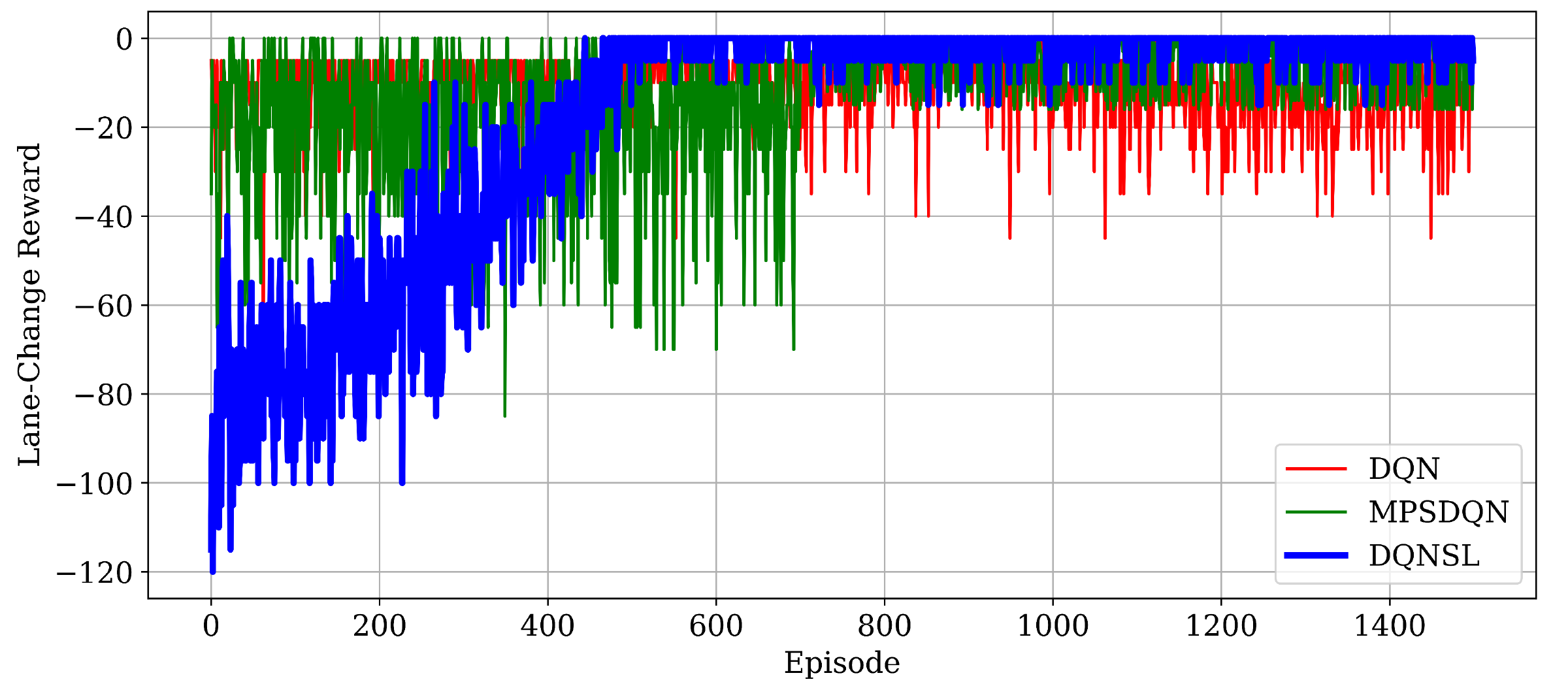}
			\caption{Lane-change reward per episode}
			\label{lanechange}
		\end{minipage}
		\hfill
		\begin{minipage}[b]{0.48\textwidth}
			\centering
			\includegraphics[width=\textwidth]{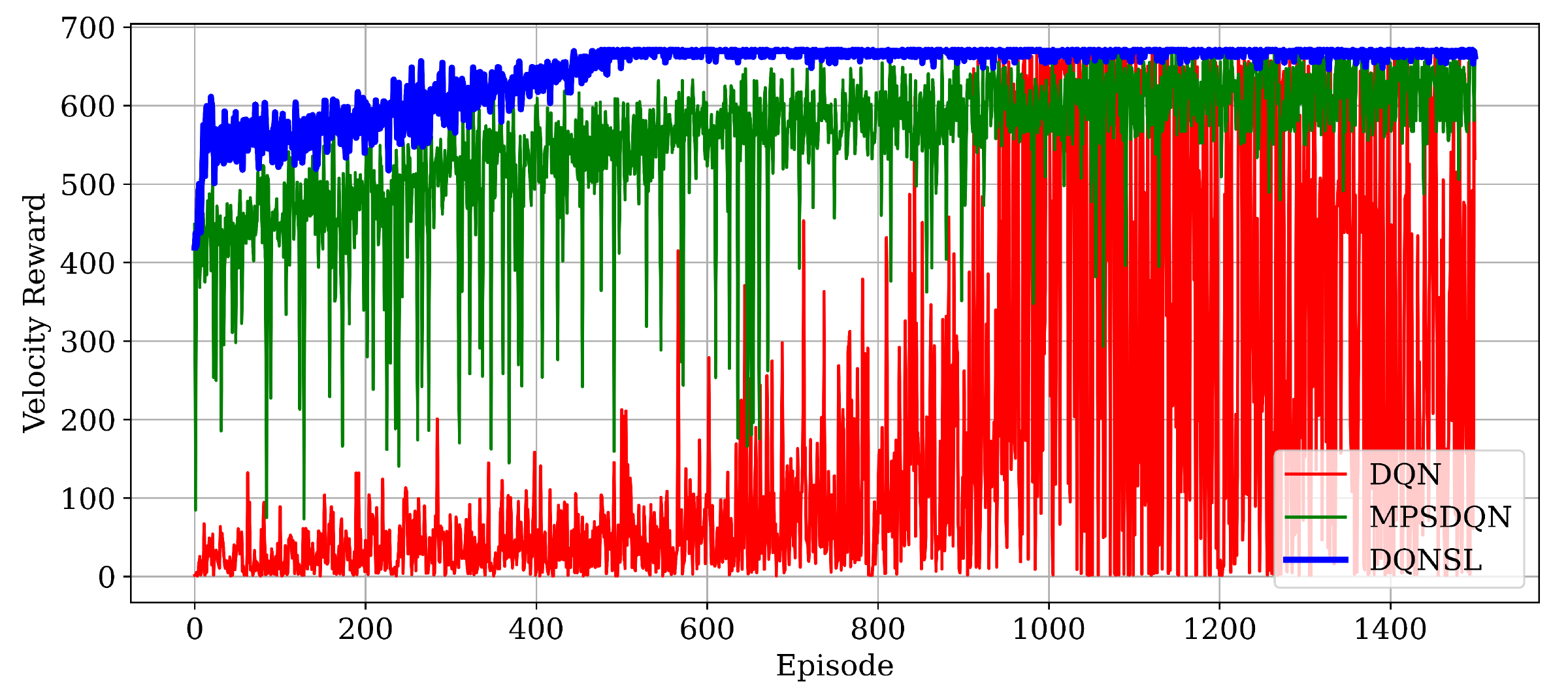}
			\caption{Velocity rewards received by agents in each episode}
			\label{velocity}
		\end{minipage}
	\end{minipage}
\end{figure*}

\subsection{Results and Discussion}
In this section, we present the results of experiments and compare them with DQN and MPSDQN. The experiments are divided to two main parts; training and test. We discuss the convergence of the reward function and the comparison of rewards between our method and other methods. Additionally, we analyze the implications of our approach in terms of safety, rule adherence, and training efficiency. Then, we test the extracted DRL models in different scenarios and evaluate each model performance.
\begin{table*}[t]
	\caption{Test results for $50$ episodes in HighD (both directions) and NGSim datasets}
	\label{test}
	\vskip 0.10in
	\begin{center}
		\begin{small}
				\begin{tabular}{lcccccccccr}
					\toprule
					DIRECTION & \multicolumn{3}{c}{Left-to-Right (HighD)} & \multicolumn{3}{c}{Right-to-Left (HighD)} & \multicolumn{3}{c}{NGSim} \\
					
					\cmidrule(r){1-1} \cmidrule(lr){2-4} \cmidrule(lr){5-7} \cmidrule(l){8-10}
					
					METHOD & \textbf{DQN} & \textbf{MPSDQN} & \textbf{DQNSL} & \textbf{DQN} & \textbf{MPSDQN} & \textbf{DQNSL} & \textbf{DQN} & \textbf{MPSDQN} & \textbf{DQNSL} \\
					
					\toprule
					
					Lane Changes & $204$  & $56$ & $43$ & $115$ & $52$  & $48$ & $91$ & $15$ & $13$ \\
					
					\cmidrule(r){1-1} \cmidrule(lr){2-4} \cmidrule(lr){5-7} \cmidrule(l){8-10}
					
					Collisions & $14$ & $4$ & $1$ & $5$ & $6$  & $2$ & $9$ & $5$ & $2$ \\
					
					\cmidrule(r){1-1} \cmidrule(lr){2-4} \cmidrule(lr){5-7} \cmidrule(l){8-10}
					
					Off-Road & $0$ & $0$ & $0$ & $33$ & $0$ & $0$ & $25$ & $0$ & $0$ \\
					
					\cmidrule(r){1-1} \cmidrule(lr){2-4} \cmidrule(lr){5-7} \cmidrule(l){8-10}
					
					Time Step & $657$ & $607$ & $592$ & $684$ & $619$ & $603$ & $971$ & $877$ & $852$ \\
					\toprule
				\end{tabular}
		\end{small}
	\end{center}
	\vskip -0.1in
\end{table*}

\textbf{Training. }
We trained all methods for $1500$ episodes using similar parameters for the reward function and the DQN network. Using the velocity control scheme, each agent was tasked with driving along an $840$-meter track to reach the end of each episode. If a collision occurred or the agent deviated from the legal lanes, the episode was terminated. The results, as shown in Figure~\ref{cumulative}, reveal that the DQNSL and MPSDQN agents learn efficiently, while the DQN agent's learning is highly unstable. This instability can be attributed to the fact that the DQN agent selects actions randomly from the entire action space, whereas other agents select actions only from the safe action set.

Furthermore, the reward results have multiple implications, particularly in terms of safety and reliability. The DQNSL and MPSDQN agents consistently receive bigger rewards, indicating that they avoid unsafe actions such as collisions, lane violations, and dangerous maneuvers. On the other hand, during the initial stages of training, the DQN agent gets significant negative rewards due to frequent lane violations and collisions. Since the behavior of other vehicles is not known for future steps, MPSDQN agent sometimes collide with other vehicles, which is the primary disadvantage of this method.

Another notable difference is the speed of training. As observed in Figure~\ref{cumulative}, the reward for the DQNSL and MPSDQN agents converge after $500$ episodes, whereas the DQN agent's reward converges in an unstable manner after $970$ episodes. The removal of unsafe actions plays a crucial role in this disparity. By constraining the action space and eliminating dangerous actions, we create a more favorable environment for the agent to discover the optimal policy. Restricting the dimensions of the action space simplifies the process of finding the best policy.

As mentioned in earlier sections, the reward function comprises four sub-reward functions: collision, off-road, lane change, and velocity rewards. Figure~\ref{collision} demonstrates that the DQNSL agent receives no penalties for collisions, indicating its successful avoidance of such incidents. Conversely, the DQN and MPSDQN agents get average penalties of $-10.65$ and $-2.26$ per episode, respectively. A similar trend can be observed for the off-road reward, as shown in Figure~\ref{outofhighway}. The DQN agent receives an average penalty of $-60.1$, while the DQNSL agent receives no penalties since it remains within the legal lanes.

The lane change and velocity rewards are the most important components of the DQNSL and MPSDQN agents' reward functions, as the other rewards remain consistently at zero. As depicted in Figure~\ref{lanechange} and Figure~\ref{velocity}, the lane change reward and velocity exhibit convergence similar to the overall reward function, reaching optimal values at the same number of episodes. Similar to MPSDQN, at the beginning of the learning process, the DQNSL agent receives negative rewards for lane changes as it explores safe state-action pairs by changing lanes. However, it gradually learns to avoid frequent lane changes and find a smoother trajectory. While the lane change reward continues to increase throughout the learning episodes, its negative value remains larger than that of the DQN agent. Since the DQN agent terminates episodes more quickly than the DQNSL agent, it does not have sufficient opportunities to engage in long maneuvers or lane changes.

Also, DQNSL and MPSDQN agents optimize the velocity reward and reaches the maximum threshold within a given episode, as observed in Figure~\ref{velocity}. The DQNSL agent's ability to regulate velocity enables it to achieve higher rewards compared to the DQN agent. The DQN agent, terminating episodes earlier, does not have the chance to fine-tune its velocity and consistently receives lower rewards.

Overall, the performance of DQNSL looks like that of MPSDQN. However, the proposed method is model-free, but knowledge-based, while MPSDQN performance depends on the accuracy of systems model. The results demonstrate that the DQNSL approach significantly enhances safety during the exploration phase and outperforms the traditional DQN agent by avoiding unsafe actions and converging more quickly. The filtering of unsafe actions using SLP ensures a higher level of safety and facilitates a faster learning process for AVs in complex environments.

\textbf{Evaluation. }
In the test scenario, we conducted 50 episodes with a track length similar to the training scenario using the HighD dataset. However, unlike training, the test scenario featured a varying number of vehicles on the highway in each episode. Additionally, we deliberately excluded the safety module during testing to ensure fair comparison of results. According to Table~\ref{test}, the findings demonstrate that all agents successfully learned to drive within the legal lanes. However, the DQN agent exhibited a significantly higher lane change frequency, averaging 5 lane changes per episode, which poses safety risks for the passengers. Also, this number is larger for MPSDQN compared to DQN. The DQNSL model limited lane changes to an average of one time per episode, leading to a smoother driving experience while avoiding unnecessary lane changes.

To compare the results of the velocity reward in the test scenarios, we assessed the performance of each agent by measuring the number of time steps required to reach the end of each track. A shorter time indicates a higher velocity reward. Table~\ref{test} shows that the DQNSL agent achieved this goal in $598$ time steps on average, whereas the DQN and MPSDQN agents required $670$ and $613$ time steps, respectively. The DQNSL agent outperformed other agents, showcasing its efficiency in completing the task within a shorter time period.
\begin{wrapfigure}{r}{0.6\textwidth}
	\centering
	\includegraphics[width=\linewidth]{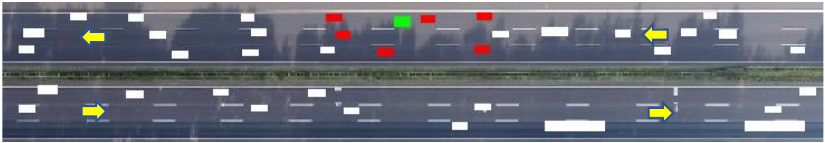}
	\caption{HighD test scenario (right-to-left direction).}
	\label{test_highD}
\end{wrapfigure}

We also examined the incidence of collisions across all test episodes. As indicated in Table~\ref{test}, the results revealed that the DQN and MPSDQN agents collided with other vehicles on the road at rates of $28$\% and $10$\%, respectively, while the collision rate for the DQNSL agent was reduced to $4$\%. It is worth noting that all collisions involving the DQNSL agent were due to the blame of other human agents, as they were unaware of the AV's presence, thereby forcing it to involve in accidents.

\begin{wrapfigure}{r}{0.5\textwidth}
	\centering
	\includegraphics[width=\linewidth]{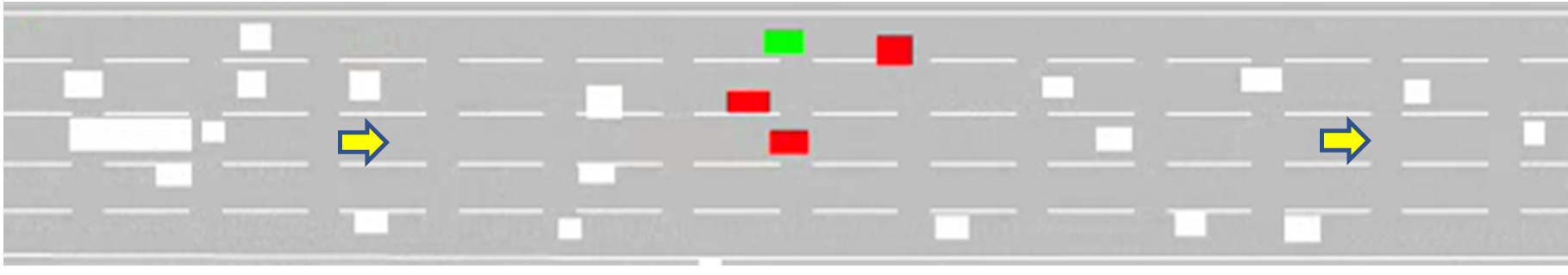}
	\caption{NGSim US-$101$ test scenario (left-to-right direction).}
	\label{test_ngsim}
\end{wrapfigure}

Furthermore, we assessed the performance of models when the AV traveled in the right-to-left direction of the HighD dataset, as shown in Figure~\ref{test_highD}, in order to evaluate the generalizability of the defined rules on the highway. According to Table~\ref{test}, the DQN agent struggled in this scenario, frequently driving off the highway. In contrast, the DQNSL and MPSDQN agents consistently remained within the boundaries of the highway. While the MPSDQN remained in the highway in the right-to-left direction, the rate of collisions increased more as compared to the DQNSL method, highlighting the effectiveness of the symbolic rules defined in the SLP core. The symbolic nature of these rules allows for easy generalization to new environments, enabling the seamless transfer and utilization of the defined rules in similar environments, ensuring broader applicability. To evaluate the generalizability of methods, we implemented them in the NGSim US-$101$ dataset, recorded in the Hollywood Freeway, Los Angeles, as shown in Figure~\ref{test_ngsim}. Though it consists of a highway scenario, the environment setting is different from that of the HighD dataset. The number of lanes is $5$, average velocity is lower than that of the HighD, and traffic density is more as compared to HighD. As shown in Table~\ref{test}, the DQNSL agent demonstrated consistent performance with minimal collisions and a reduced number of lane changes, achieving smoother driving behavior and completing episodes in fewer time steps. Notably, this was achieved without retraining or adapting the symbolic rules, underscoring their transferability to diverse highway environments.

Overall, we have improved the safety of the DQN agent during the exploitation phase as well by eliminating dangerous actions from the set of possible actions during this phase. This enhancement brings DRL algorithms closer to the realities of AD systems. Moreover, we have shown that our method is able to perform well in the right-to-left HighD driving scenario as well as the NGSim environment, indicating the generalizability of the symbolic rules which improves the transferability of the model to new highway environments.

\section*{Conclusion and Future Research}
In conclusion, this article proposed a novel approach for model-free deep reinforcement learning with symbolic logic (DRLSL) for safe autonomous driving. The main contribution of this approach is to ensure safety during the exploration phase of DRL by filtering out unsafe actions from the action set in a given state. The proposed method was evaluated on the HighD and NGSim datasets, and the results showed a significant improvement in the safety level of the DRL algorithm. We trained the autonomous vehicle agent without significant collisions, assisting the DRL agents to come to reality when training. Moreover, we significantly improved the stability and convergence speed of DRL during training. The result of the test scenarios indicated that our method is capable of driving with far less collision rates and lane violations compared to DRL agent and model predictive shielding with DRL agent. Moreover, in contrast to DRL only, DRLSL was able to drive in other side of the highway and NGSim as another different environment, placing emphasis on the generalizability of our algorithm. This approach has great potential to enhance the safety of autonomous driving systems, and it can be extended to other domains of DRL where safety is a critical concern.

We encourage future research to implement DRLSL in different driving scenarios (e.g. vehicle turns, pedestrian interventions, intersections, etc.) by finding safety rules specific to each scenario. Moreover, this methodology can be challenged when used in real-world applications with uncertain sensory information or more difficult tasks like overnight driving. Another potential direction is to extract new rules using examples (e.g. inductive logic programming) \citep{sharifi2023symbolic} or real-time experiences in order to make the learning process more self-sufficient.





\bibliography{iclr2025_conference}
\bibliographystyle{iclr2025_conference}
\end{document}